%% file: main.tex
\documentclass[sigconf]{acmart}

\renewcommand\footnotetextcopyrightpermission[1]{} 
\usepackage{booktabs} 
\usepackage{multirow}
\usepackage{amsmath}
\usepackage{mathtools}

\newcommand{\xhdr}[1]{\vspace{2mm}\noindent{{\bf #1}}}

\usepackage{bm}
\renewcommand{\vec}[1]{\bm{#1}}

\setcopyright{none}

\acmDOI{}

\acmISBN{}

\acmConference[AI for Fashion]{AI for Fashion: KDD 2018}{August 2018}{London} 
\acmYear{2018}
\copyrightyear{2018}

\begin{document}
\title{Visually-Aware Personalized Recommendation using Interpretable Image Representations}


\author{Charles Packer}
\affiliation{%
  \institution{University of California, San Diego}
  \city{La Jolla} 
  \state{California} 
  \country{USA}
}
\email{cpacker@cs.ucsd.edu}

\author{Julian McAuley}
\affiliation{%
  \institution{University of California, San Diego}
  \city{La Jolla} 
  \state{California} 
  \country{USA}
}
\email{jmcauley@cs.ucsd.edu}

\author{Arnau Ramisa}
\affiliation{%
  \institution{Wide Eyes Technologies}
  \city{Barcelona} 
  \country{Spain} 
}
\email{aramisa@wide-eyes.it}

\begin{abstract}
\emph{Visually-aware} recommender systems use visual signals present in the underlying data to model the visual characteristics of items and users' preferences towards them. 
In the domain of clothing recommendation, incorporating items' visual information (e.g., product images) is particularly important since clothing item appearance is often a critical factor in influencing the user's purchasing decisions. 
Current state-of-the-art visually-aware recommender systems utilize image features extracted from pre-trained deep convolutional neural networks, however these extremely high-dimensional  representations are difficult to interpret, especially in relation to the relatively low number of visual properties that may guide users' decisions. 

In this paper we propose a novel approach to personalized clothing recommendation that models the dynamics of individual users' visual preferences. 
By using interpretable image representations generated with a unique feature learning process, our model learns to explain users' prior feedback in terms of their affinity towards specific visual attributes and styles.
Our approach achieves state-of-the-art performance on personalized ranking tasks, and the incorporation of interpretable visual features allows for powerful model introspection, which we demonstrate by using an interactive recommendation algorithm and visualizing the rise and fall of fashion trends over time.
\end{abstract}



\maketitle

\input{body}

\bibliographystyle{ACM-Reference-Format}
\bibliography{ref} 

\end{document}

%% file: body.tex
\section{Introduction}


Recommender systems help to discover items of personal interest by learning from historical feedback in order to understand the factors that influence users' decisions. Recently, there has been an interest in developing recommender systems that are `visually aware,' in the sense that the visual features (extracted from product images) are incorporated directly into the recommendation objective. Such systems can substantially improve recommendation accuracy, especially in settings (such as clothing recommendation) where visual factors strongly guide users' decisions.

However, actually incorporating visual signals can be challenging. Extracting meaningful representations (the complexity of style) from image data alone is not straightforward, and can require costly, high-dimensional representations (e.g.~CNN-based methods). Furthermore, high-dimensional `black box' image models offer little by way of interpretability, which can impede usability when building interfaces that interact with these representations.

In this paper we seek to build visually-aware representations on top of \emph{interpretable} visual features based on fine-grained parsing of product images, for the problem of clothing recommendation. We show that such features can lead to superior performance compared to `black box' image representations, while substantially reducing their dimensionality, and also that such features can be used to develop more usable and interactive systems.




\section{Related Work}

We build upon latent factor models, and in particular Bayesian Personalized Ranking (BPR) \cite{rendle2009bpr}, which is trained using \emph{implicit feedback} (i.e., purchases vs.~non-purchases) in order to estimate rankings of items that are likely to
be
interacted with. In particular, our work extends ideas from visually aware recommendation as well as models of fashion and clothing style.

\xhdr{Visually-aware Recommender Systems.}

Recent works have introduced visually-aware recommender systems where users' rating dimensions are modeled in terms of visual signals in the system (product images). 
Systems have been built for link prediction \cite{julian} and personalized search,
though most closely related are methods that extend 
traditional recommender systems (such as Bayesian Personalized Ranking)
to incorporate
visual dimensions to facilitate item recommendation tasks \cite{HeMcA16VBPR}. We build on extensions to such models that incorporate temporal dynamics in addition to visual signals to capture the evolution of fashion style \cite{HeMcA16aTVBPR}.

\xhdr{Fashion and Clothing Style.}
Beyond the methods mentioned above, modeling fashion or style characteristics has emerged as a popular computer vision task in settings other than recommendation,
e.g.~with a goal to
categorize or extract features from images, without necessarily building any model of a `user.'
This includes categorizing images as belonging to a certain style \cite{bossard2013apparel}, as well as models that create rich stylistic annotations, like \emph{DeepFashion} \cite{liuLQWTcvpr16DeepFashion}.



\section{Model}

\subsection{Visual Feature Generation} \label{sec:pipeline}

To compute the features and attribute probabilities used in the recommendation experiments, we have implemented a variation of the model proposed in~\cite{Dong2016a}, without including the cross-domain transfer learning architecture, as it was not deemed necessary for our dataset. The network was initially trained using the ImageNet dataset \cite{imagenet}, to obtain a general set of intermediate feature representations, and subsequently fine-tuned for several epochs on a large proprietary dataset of fashion images annotated with the target attributes. Care was put into making sure that each attribute is represented to a sufficient extent in the fine-tunning dataset, to guarantee a consistent degree of generalization to new types of images for all classes. 

In terms of performance, our model does comparably or better than DeepFasion~\cite{liuLQWTcvpr16DeepFashion} or MTCT~\cite{Dong2016a} according to the numbers reported in their respective papers, although it is difficult to compare exactly as the test datasets are different, and our attribute categories are not necessarily the same as theirs.

{\small
\begin{table}[t]
  \caption{Notation.}
  \label{table:notation}
  \begin{tabular}{ll}
    \toprule
Notation & Definition\\ 
	\midrule
$\mathcal{U},\mathcal{I}$ & user set, item set \\
$\hat x_{u,i}$ & predicted `score' user $u$ gives to item $i$ \\
$K$ & number of latent factors \\
$K^\prime$ & number of visual factors \\
$F$ & number of image features \\
$\alpha$ & global offset (scalar) \\
$\beta_u, \beta_i$ & bias of user $u$, item $i$ (scalar) \\
$\gamma_u, \gamma_i$ & latent factors of user $u$, item $i$ ($K \times 1$) \\
$\theta_u, \theta_i$ & visual factors of user $u$, item $i$ ($K^\prime \times 1$) \\
$f_i$ & visual (image) features of item $i$ ($F \times 1$) \\
$\mathbf{E}$ & embedding matrix ($K^\prime \times F$) \\
$\beta$ & visual bias vector ($F \times 1$) \\
$\vec{p}_u$ & user $u$'s affinity vector ($F \times 1$) \\
$\vec{p}_u(t)$ & user $u$'s affinity vector at time $t$ ($F \times 1$) \\
$\vec{p}_{u,k}$ & $k$'th index of user $u$'s affinity vector (scalar) \\
$\vec{h}(k)$ & one-hot vector `on' at index $k$ ($F \times 1$) \\
$\phi$ & item scaling factor (scalar) \\
$\phi^\prime$ & feature scaling factor (scalar) \\
    \bottomrule
  \end{tabular}
\end{table}
}

\subsection{Bayesian Personalized Ranking}

The core of our prediction model is built on Matrix Factorization (MF), a state-of-the-art method for rating prediction. 
The basic MF formulation describes each user's preference towards an item in terms of a set of user and item specific latent factors  ($\gamma_u$, $\gamma_i$), such that the inner product $\gamma_u^T \gamma_i$ encodes the compatibility between $u$ and $i$. 
In our case, the preference predictor extends the basic latent factor model by learning set of visual factors ($\theta_u$, $\theta_i$), where $\theta_u^T \theta_i$ encodes a separate visual-specific compatibility. 
Using the image features directly for $\theta_i$ is problematic due to the high dimensionality of $f_i$, so we learn an embedding kernel $\mathbf{E}$ which maps our image features to a lower dimensional space ($\theta_i = \mathbf{E}f_i$). 
Thus, a user $u$'s predicted rating of an item $i$ is given by the following predictor:
\begin{equation}
\hat x_{u,i} = 
\underbrace{
\alpha + 
\beta_u + 
\beta_i + 
\beta^{T}f_i
}_{\mathclap{\text{bias terms}}}
+ 
\underbrace{
\gamma_u^T\gamma_i + \theta_u^T(\mathbf{E}f_i)
}_{\mathclap{\text{user-item interactions}}}
\label{eq:pred_theta_fac}
\end{equation}






\subsubsection{Temporal Dynamics}
To incorporate temporal dynamics in addition to visual signals, we employ the technique introduced in \cite{HeMcA16aTVBPR} which extends eq. \ref{eq:pred_theta_fac} by parameterizing the visual factors bias by a set of learned epochs (a fixed number of flexible time-based dataset partitions). Critically, we can describe an items' visual factors at time (epoch) $t$ in terms of a time-specific weighting vector $w(t)$ (users weigh  visual dimensions, or `styles', differently over time) and temporal drift (items gain and lose attractiveness in different dimensions over time):
\begin{equation}
\theta_i(t) = 
\underbrace{
\mathbf{E}f_i \odot w(t) 
}_{\mathclap{\text{base}}}
~~~+~~~
\underbrace{
\bigtriangleup_{\mathbf{E}}(t) f_i
}_{\mathclap{\text{temporal drift}}}
\label{eq:thetai}
\end{equation}
where $\odot$ indicates the Hadamard product.

\subsubsection{Fitting the Model}

To fit our model, we use Bayesian Personalized Ranking (BPR), a pairwise ranking optimization framework. BPR adopts stochastic gradient descent to efficiently learn the parameters of the model, using the desired preference predictor and implicit feedback from the training data. The same training procedure is used for the temporally-aware model, adjusted to incorporate review timestamps as additional feedback (used to learn the optimal epoch segmentation). 
For further details regarding the training procedure and model formulation, 
refer to \cite{HeMcA16aTVBPR}. 

\section{Experiments}

\begin{table*}[t]
  \caption{
AUC on the test set. 
Boldface indicates the models using interpretable visual features. 
All models' hyperparameters were optimized during training for AUC on the validation set.  
\emph{Women} and \emph{Men} are the two largest subcategories within \emph{Clothing$^\mathit{+}$}.
}
  \label{table:auc}
  \begin{tabular}{llcccccccc}
    \toprule


    Dataset &
    Setting &
    RAND &
    POP &
    MM-MF &
    BPR-MF &
    VBPR &
    \textbf{I-VBPR} &
    TVBPR &
    \textbf{I-TVBPR} 
     \\
     
    \midrule
    
    \multirow{2}{*}{\emph{Clothing$^\mathit{+}$ (all)}} &
    All Items &
    0.504556 & 
    0.443571 & 
    0.640032 & 
    0.645513 & 
    0.748643 & 
    0.744377 & 
	0.761767 &
    0.749785  
    \\
    & \emph{Cold Start} &
    0.505772 & 
    0.267167 & 
    0.529043 & 
    0.537888 & 
    0.699001 & 
    0.688506 & 
	0.731984 &
    0.691508   
    \\
    
    \addlinespace[0.3em]
    \multirow{2}{*}{\emph{Women's clothing}} &
    All Items 
    & 0.494471 
    & 0.391156 
    & 0.612812 
    & 0.621887 
    & 0.741344 
    & 0.740134 
    & 0.758179 
    & 0.745743 
    \\
    & \emph{Cold Start}
    & 0.501157 
    & 0.249498 
    & 0.524881 
    & 0.537444 
    & 0.698104 
    & 0.705267 
    & 0.716321 
    & 0.707428 
    \\

    \addlinespace[0.3em]
    \multirow{2}{*}{\emph{Men's clothing}} &
    All Items
    & 0.505361 
    & 0.356822 
    & 0.629649 
    & 0.631144 
    & 0.714754 
    & 0.726443 
    & 0.734160 
    & 0.730461 
    \\
    & \emph{Cold Start}
    & 0.494187 
    & 0.236677 
    & 0.565715 
    & 0.572270 
    & 0.669684 
    & 0.701436 
    & 0.702629 
    & 0.704644 
    \\

 
    \bottomrule
  \end{tabular}
\end{table*}

\subsection{Dataset}

We use a modified version of the \emph{Amazon.com} dataset introduced by McAuley et al \cite{HeMcA16aTVBPR}. The original dataset is a content-rich dataset containing millions of items, metadata, images, and implicit-feedback (e.g., review data), which we use for our ground truth `preference' statistic. 

In our testing, we use a subset of the original dataset (\emph{Clothing$^+$}) containing 1.4 million items filtered from categories that encode fashion dynamics (clothing, jewelry, bags, etc). Additionally, we report results on the largest two subcategories within our dataset, \emph{Women} and \emph{Men} (643,195 and 278,762 items, respectively). As part of preprocessing, we filter out users who have less than 5 written reviews to increase the density of the dataset. 

\subsection{Evaluation Methodology}

Our data is split into training, validation and test sets by sampling one item for validation ($\mathcal{V}_u$) and another for testing ($\mathcal{T}_u$) for each user, and the remaining data is used for training ($\mathcal{P}_u$). 
Similar to \cite{HeMcA16aTVBPR}, all methods reported are evaluated on $\mathcal{T}_u$ with the widely used AUC (Area Under the ROC curve) metric: 
\begin{equation}
AUC = 
\frac{1}{|\mathcal{U}|} 
\sum_u 
\frac{1}{|E(u)|}
\sum_{(i,j) \in E(u)}
\delta(\hat x_{u,i} > x_{u,j})
\label{eq:auc}
\end{equation}

\noindent where $\delta(b)$ is an indicator function that returns 1 iff $b$ is true and the set of evaluation pairs for user u is: 
\begin{equation}
E(u) = {(i, j)|(u, i) \in \mathcal{T}_u \wedge (u, j) \notin/ (\mathcal{T}_u \cup \mathcal{V}_u) \cup \mathcal{P}_u}
\label{eq:e(u)}
\end{equation}

All methods are evaluated under two settings, `All Items' and `Cold Start'. `All Items' evaluates the average AUC across the entire test set, whereas `Cold Start' evaluates the average AUC for items with less than five recorded feedback instances in the training set. Cold start performance is particularly important in the clothing recommendation setting, where most datasets will have long-tailed distributions due to the constant flow of new items with no prior feedback. 

\subsection{Comparison Methods}

We compare our method using low-dimensional interpretable visual features (I-VBPR) and additional temporal dynamics (I-TVBPR) with state-of-the-art visually- and temporally-aware methods (VBPR and TVBPR) using `black box' image features extracted from AlexNet \cite{alexnet}. 
We also compare our method to several Matrix Factorization approaches for reference (both Matrix Factorization-based baselines (\textsc{MM-MF} and \textsc{BPR-MF}) were implemented using MyMediaLite\footnote{\url{http://www.mymedialite.net}}). MM-MF is a pairwise MF model optimized for hinge ranking loss, while BPR-MF is the state-of-the-art (non-visual method) for personalized ranking on implicit feedback datasets. RAND (random) assigns preferences at random, while POP (popularity) rank prediction is equivalent to an item's popularity.


\subsection{Performance and Analysis}

Table \ref{table:auc} shows the average AUC for each method on the \emph{Women's}, \emph{Men's}, and overall \emph{Clothing$^+$} test datasets. 
We used 10 latent factors for all models, and 10 additional visual factors for the visually-aware models. 
For the visually-aware methods, $\lambda_{\theta}$ was set to 5 and the remaining regularization hyperparameters were set to 0. 
We summarize the results as follows:

\xhdr{Temporally and visually-aware methods.} Incorporating visual signals clearly significantly increases performance, and all visually-aware methods improve at least 10\% compared to BPR-MF, the next-best baseline. 
Incorporating temporal information information on top of visual information increased accuracy in all cases, though with smaller margins. 


\xhdr{Interpretable vs. `black box' image features.} 
Due to the difference in dimensionality of $f_i$ between the interpretable features and black box features (several hundred vs. several thousand image features), the interpretable models use a small fraction of the total model parameters compared to the black box models (<5\%). 
Despite the large discrepancy in parameter count, models using the interpretable features achieve comparable prediction accuracy: in the overall dataset and in \emph{Women's Clothing}, the black box methods maintained a 1-2\% performance boost in the `All Items' setting, while the interpretable methods maintained a similar lead in most of \emph{Men's Clothing} and several cold start scenarios. 


Ultimately, our results demonstrate that by using a relatively low number of interpretable image features, our model can produce comparable and in several cases superior results to the black box approach while substantially reducing model complexity and allowing for more usable and interactive systems. 


\section{Interactive recommendation}

In this section we demonstrate how our prediction model can be extended to build a personalized, interactive, \emph{fashion-aware} recommendation system. 
In addition to improving raw performance, another tangible benefit of incorporating interpretable visual features into our prediction model is that it allows us to model users' preferences in terms of directly observable item properties. 
Unlike the typical features extracted from pre-trained CNNs, interpretable visual features allows the recommender system to generate sets of recommendations due to a specific feature, or similarly, to narrow down a set of items based on explicit visual criteria.



\subsection{User personalization}

We initialize the recommender system for user $u$ by constructing an \emph{affinity vector} $\vec{p}_u$, which represents the sensitivity of user $u$ towards each visual dimension. This is done by first fitting the model, then recording an average response for each visual dimension using a modified version of the preference predictor function (eq. \ref{eq:pred_theta_mod}). Given a user $u$ and a feature dimension $k$, we calculate an average the response towards $k$ using a one-hot vector (1.0 at index $k$, 0.0 at all other indices) across a random sample of items $R(\mathcal{I})$ and store the result in the affinity vector:  
\begin{equation}
\underbrace{\vec{p}_{u,k}(t=0)}_{\mathclap{\text{affinity vector}}} = \sum_{i \in R(\mathcal{I})} \underbrace{\alpha + \beta_u + \beta_i + \gamma_u^T\gamma_i + \theta_u^T(\mathbf{E} \vec{h}(k)) + \beta^{T}\vec{h}(k)}_{\mathclap{\text{response of $u$ using one-hot visual encoding}}}
\label{eq:pred_theta_mod}
\end{equation}

Once the response towards each dimension $k$ has been recorded in the respective $\vec{p}_{u,k}$, we rescale the affinity vector to match the original feature scaling using a normalization function $\mathbf{N}(\cdot)$ that divides each element by the sum of the elements in $\vec{p}_{u}$.
This allows us to uncover which visual dimensions the user is most responsive to.



\subsection{User interaction}

\begin{figure}[t]
\includegraphics[width=\columnwidth]{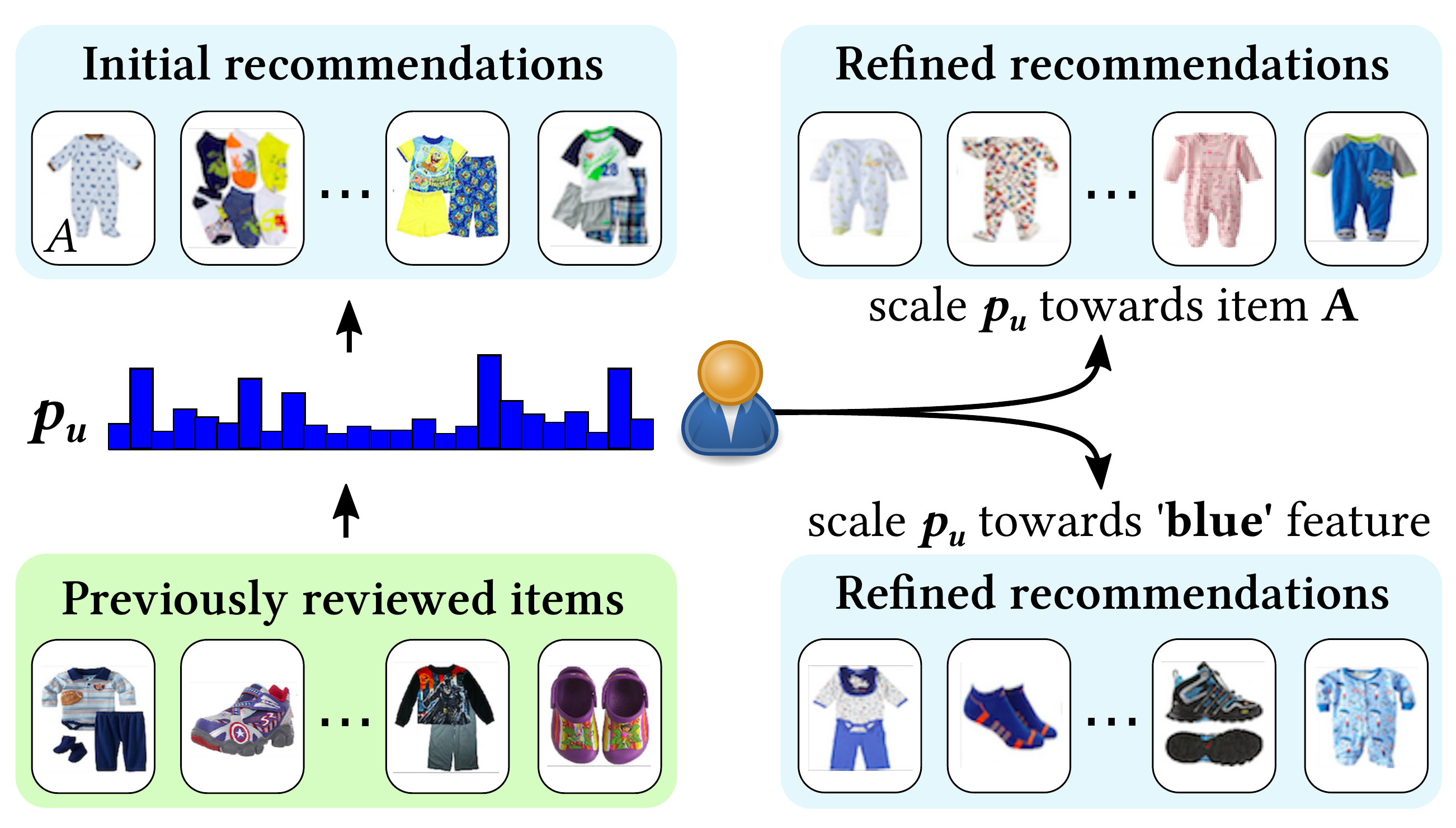}
\caption{Interactive personalized recommendation system. A user's affinity vector $\vec{p}_u$ is initialized based on prior feedback. A user can then tailor their recommendations to be more like a specific item or feature, modifying $\vec{p}_u$.}\label{fig:demo}
\end{figure}

Our goal is to not only generate highly-personalized item recommendations based on visual signals, but also to allow the user to tailor their own recommendation results dynamically. 
Once the affinity vector has been initialized, we can begin to generate item recommendations using a nearest neighbor search within the item set using the visual feature space. A user can dynamically update the model's generated recommendations by scaling their own affinity vector in the direction of a chosen item (see figure \ref{fig:demo}):

\begin{equation}
\vec{p}_u(t) = 
\mathbf{N}(
\underbrace{
\vec{p}_{i}(t) + \phi \cdot \vec{f}_{i}
}_{\mathclap{\text{scale $\vec{p}_u$ towards item $i$}}}
)
\label{eq:updateitem}
\end{equation}
\noindent Additionally, a user can choose to boost a specific visual feature (e.g., color) within their affinity vector: 
\begin{equation}
\vec{p}_u(t) = 
\mathbf{N}(
\underbrace{
\vec{p}_{u,k}(t) + \phi^\prime \cdot \vec{h}(k)}_{\mathclap{\text{scale $\vec{p}_u$ towards visual dimension $k$}}}
)
\label{eq:updatefeature}
\end{equation}


\noindent In each case, the scaling constants $\phi$ and $\phi^\prime$ are fixed prior to runtime and determined experimentally.

\section{Tracking fashion trends}

Previous work has focused on visualizing the temporal shift of the latent visual dimensions, by plotting the weighting vector $w(t)$ (eq. \ref{eq:thetai}) at each epoch \cite{HeMcA16aTVBPR}. However, the meaningfulness of such visualizations is ambiguous, since it requires inferring the visual property or style (among many) the latent dimension may be capturing. 


Since our model utilizes interpretable visual features (instead of extracted CNN features), we are able to visualize the temporal-dynamics of our dataset at the feature-level. To track the popularity of a feature within a learned epoch $t$, we can sum the time-weighted \emph{influence} of a vector on each latent visual dimension:

\begin{equation}
\sum_{K^\prime} 
\underbrace{
\mathbf{E}\vec{h}(k) \odot w(t)
}_{\mathclap{\text{weighted influence of feature $k$ at epoch $t$}}}
\label{eq:vis}
\end{equation}

\begin{figure}[t]
\includegraphics[width=\columnwidth]{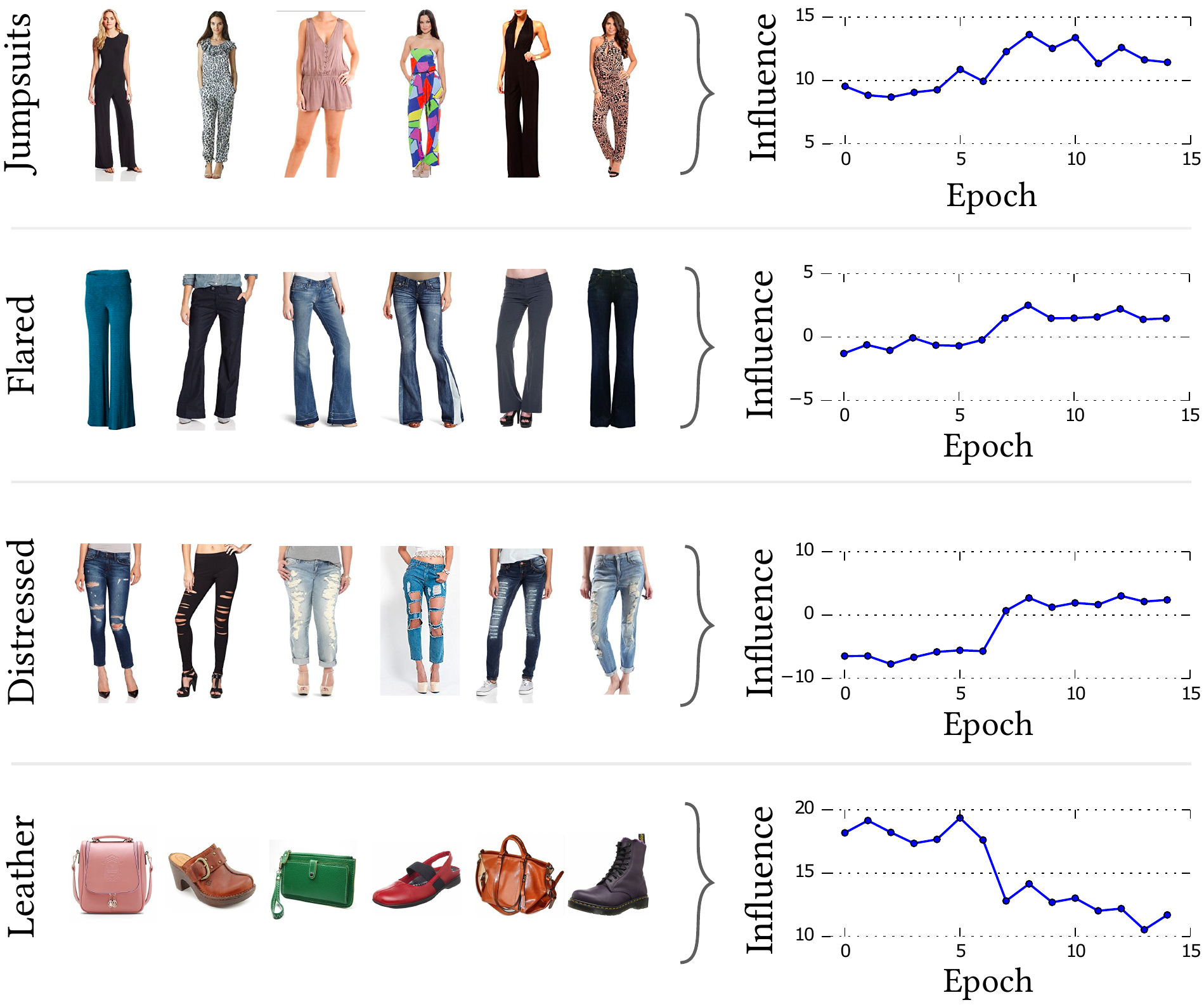}
\caption{Tracking fashion trends throughout time using interpretable visual features. Our feature generation architecture learns  variety of categorical (`jumpsuits'), material-based (`leather'), and style-based (`distressed', `flared') attributes. Selected images are items from $R(\mathcal{I})$ with the highest loss at the selected feature ($\max_i f_{i,k}$).  To track the popularity of individual features, we observe a feature's influence ($\sum_{K^\prime} \mathbf{E}\vec{h}(k) \odot w(t)$) across each epoch.}\label{fig:trends}
\end{figure}

Thus, given an interpretable feature $k$, using a feature $k$'s influence we can model how the feature's popularity has changed over time (see figure \ref{fig:trends}). This allows us to directly observe 
the evolution of real interpretable feature dimensions, 
as opposed to the ambiguous dimensions of the latent embedding.



\section{Conclusion}


In this paper we introduced a novel approach to fashion-aware product recommendation that utilizes interpretable visual features. 
We show that such features can lead to superior performance compared to `black box' image representations, while substantially reducing their dimensionality, and also that such features can be used to develop more usable and interactive systems.
Future work will focus on extended applications enabled by the feature generation process, 
including iterative clustering and querying of recommendation results. 


